\documentclass{article} 
\usepackage[preprint]{colm2026_conference}
\usepackage[pdftex]{graphicx}
\usepackage{microtype}
\usepackage{hyperref}
\usepackage{url}
\usepackage{booktabs}
\usepackage{amsthm}
\usepackage{amsmath}
\usepackage{multirow}
\usepackage{adjustbox}

\usepackage{lineno}

\definecolor{darkblue}{rgb}{0, 0, 0.5}
\hypersetup{colorlinks=true, citecolor=darkblue, linkcolor=darkblue, urlcolor=darkblue}

\title{Know Your Limits : On the Faithfulness of LLMs as Solvers and Autoformalizers in Legal Reasoning}

\author{
  \begin{tabular}{ccc}
    \bf Olivia Peiyu Wang$^1$ & \bf Sanna Wong-Toropainen$^{2,3}$ & \bf Daneshvar Amrollahi$^4$ \\[6pt]
    \bf Ryan Bai$^5$ & \bf Tashvi Bansal$^6$ & \bf Arush Garg$^7$ \\[6pt]
    & \bf Leilani H. Gilpin$^1$ & \\[8pt]
  \end{tabular} \\[4pt]
  \small
  $^1$UC Santa Cruz \quad $^2$Univ. Helsinki \quad $^3$CodeX, Stanford \quad $^4$Stanford University \\
  $^5$Canyon Crest Academy \quad $^6$Monta Vista High School \quad $^7$Los Altos High School
}
\begin{document}

\ifcolmsubmission
\linenumbers
\fi

\maketitle

\begin{abstract}
Large Language Models (LLMs) achieve strong performance on reasoning tasks, but whether this reflects faithful logical inference or heuristic approximation remains unclear. We study this question in legal entailment by comparing three paradigms, including pure LLM classification, LLM-based Formal Reasoning, and solver-based Formal Reasoning using the Z3 SMT solver, on a re-annotated subset of ContractNLI across five LLMs. Our re-annotation reveals a systematic and measurable gap between pragmatic legal interpretation and strict formal entailment, where a substantial proportion of legally sound inferences are not formally grounded without additional unstated assumptions. While introducing formal structure improves accuracy, with LLM-based Formal Reasoning achieving the highest benchmark performance, we show that this gain does not imply faithful reasoning. We identify three recurring failure modes: \textit{scope laundering}, where LLMs report solver-inconsistent classifications without executing the underlying formal reasoning, producing conclusions that appear logically grounded but are not; \textit{implicit constraint blindness}, where LLMs overlook logical constraints present in formal representations; and \textit{program synthesis failures}, where LLMs generate incorrect Z3 code despite structured prompting. Critically, scope laundering persists across all models, raising serious concerns about the faithfulness of LLM-based formal reasoning as a proxy for symbolic execution. These results reveal a fundamental gap between benchmark accuracy and logical faithfulness.
\end{abstract}

\section{Introduction} 

The adoption of Large Language Models (LLMs) in high-stakes domains such as law is constrained by the need for faithful and verifiable reasoning. While techniques such as Chain-of-Thought improve performance \citep{wei2022chain, chen2025reasoning}, LLMs remain prone to hallucination, fabricated citations, and unverifiable inference.  Some failures are unacceptable in legal settings \citep{dahl2024large}. Beyond hallucination, correct conclusions often depend on implicit assumptions that are not explicitly stated, making it unclear which inferences are permissible. This ambiguity contributes to the significant verification overhead required for legal deployment \citep{dixon2025guidelines}.

Neuro-symbolic approaches address this by combining LLM-based autoformalization with symbolic reasoning, translating natural language into formal representations evaluated under strict semantics. However, this exposes a key limitation: formal reasoning requires explicit assumptions, while legal text is inherently underspecified.

In this work, we compare three paradigms: pure LLM classification, LLM-based formal reasoning, and LLM-driven autoformalization with Satisfiability Modulo Theories (SMT) solving. In SMT solving, an SMT solver formally determines whether a set of first-order logic formulas is satisfiable with respect to a background theory, returning SAT or UNSAT, using SMT-LIB as its input language. We evaluate these methods on a modified ContractNLI dataset \citep{koreeda2021contractnli} of real-world legal contracts, across five LLMs: GPT-OSS-120B(GPT) \citep{agarwal2025gpt}, Claude-Sonnet-4-6(Claude) \citep{anthropic2026sonnet46}, Meta-Llama-3.1-8B(Llama) \citep{meta2024llama3}, DeepSeek-V3.2(Deepseek) \citep{liu2025deepseek}, and Qwen2.5-72B(Qwen) \citep{qwen2}.

We find that adding structure improves accuracy, with LLM-based formal reasoning performing the best. However, this does not imply faithful reasoning: LLMs misinterpret formal representations,  introduce unstated assumptions, and output plausible classification without invoking the solver on the formal logic. Meanwhile, solver-based reasoning remains conservative by enforcing strict logical validity. 

These results reveal a fundamental gap between benchmark accuracy and logical faithfulness, 

\textbf{Our contributions are as follows:}
\begin{itemize}
    \item We show that adding formal structure improves LLM performance, with LLM-based Formal Reasoning outperforming both pure LLM and neuro-symbolic pipelines.
    \item We demonstrate that improved accuracy does not imply faithful reasoning, as LLMs misinterpret formal representations and rely on implicit assumptions without executing the formal logic for their classification results.
    \item We identify three major failure modes, including scope laundering, implicit constraint blindness, and program synthesis failures, surfacing the tension between accuracy and faithfulness.
    \item We identify assumption ambiguity as a core challenge, where the boundary between valid inference and unjustified assumptions is unclear.
    \item We highlight a gap between benchmark labels and formal logical semantics in real-world legal data.
\end{itemize}

\section{Related Work}
\paragraph{Neuro-symbolic Reasoning with Autoformalization and Symbolic Solving.}
A key limitation of pure LLM-based reasoning is the lack of faithfulness: chain-of-thought outputs often reflect heuristics rather than logical inference \citep{turpin2023language, chen2025reasoning}. Structured LLM approaches, such as Selection-Inference \citep{creswell2022faithful}, impose reasoning structure but do not eliminate hallucination. Neuro-symbolic methods instead decompose reasoning into LLM-based autoformalization and symbolic execution, where the latter guarantees correctness with respect to its inputs \citep{han2022folio, yang2025neuro, ryu2024divide, yang2023harnessing}. This shifts the challenge to representation quality. However, prior work shows that LLM-generated formalizations are often incomplete or inconsistent \citep{pan2023logic, vakharia2024proslm}, raising concerns about whether implicit assumptions required for correct reasoning can be reliably captured. Another line of work explored the approach of utilizing LLM as the solver and showed that this approach outperformed others while reducing syntax errors \citep{li2024leveraging, xu2024faithful, feng2024language}. However, they did not evaluate whether the performance boost and error reduction came at the cost of faithfulness to the results generated by the symbolic solver. 
Recent work detects such formalization drift via round-trip equivalence checking \citep{amrollahi2026faithfulautoformalizationroundtripverification}, but does not study which unstated assumptions are justified. Our work assesses and compares the performance of three different modes, ranging from pure LLM classification, LLM-based formal reasoning, and LLM-driven autoformalization with Satisfiability Modulo Theories (SMT) solving, and shows that structures improve performance, but the performance boost may come at the cost of faithfulness. 

\paragraph{Neuro-symbolic AI and Formal Reasoning in the Legal Domain.}
Applications of neuro-symbolic reasoning in law remain limited and often rely on simplified or manually annotated data \citep{holzenberger2020dataset, holzenberger2021factoring, holzenberger2023connecting}. Existing approaches typically employ domain-specific languages such as Stipula \citep{hahnle2025formal} and Catala \citep{merigoux2021catala}, which provide strong guarantees but are difficult to integrate with general-purpose solvers. More broadly, formal reasoning benchmarks such as FOLIO \citep{han2022folio} and ProofWriter \citep{tafjord2021proofwriter} rely on synthetic or simplified language, limiting their applicability to real-world legal reasoning. When LLM-based approaches are used, they are typically evaluated on simplified inputs that do not reflect the ambiguity and underspecification of real contracts \citep{jurayj2026language}. A typical example from the simplified dataset looks like "All people who regularly drink coffee are dependent on caffeine." By contrast, even a short legal example reads: "Neither Party shall, without the written approval of the other Party, publish, copy, or use the Confidential Information for their sole benefit." Within the legal domain, datasets frequently simplify text or focus on narrow subfields, and rarely address the role of implicit assumptions in interpretation.  ContractNLI \citep{koreeda2021contractnli} is a notable exception, grounding entailment tasks in authentic contract language. However, its labels reflect interpretation rather than strict logical entailment, motivating our re-annotation to align the dataset with formal reasoning. The US health insurance policy dataset from \cite{kant2025towards} is another dataset that is based on real insurance policies; however, it is not publicly available. Our work starts from the only publicly available real legal contract dataset - ContractNLI, and re-annotates the labels based on the logic definitions of "entailment", "contradiction", and "neutral". Through our experiments and qualitative analysis, we surface the gaps between legal interpretation and logic solving. 

\section{Methodology}
\subsection{Dataset Construction}
\begin{figure}[t]
\vskip 0.05in
\begin{center}
     \includegraphics[width=1\linewidth]{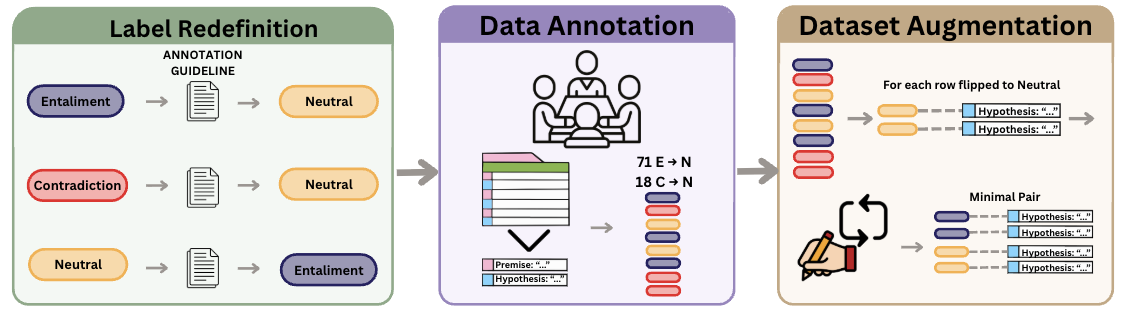}
     \caption{Visualization of the Dataset Construction Process}
     \label{fig:dataset}
\end{center}
\vskip -0.2in
\end{figure}

To evaluate reasoning under realistic legal conditions, we build on the ContractNLI dataset \citep{koreeda2021contractnli}, a benchmark dataset that was derived from 607 authentic Non-disclosure Agreements (NDAs). However, its original labels reflect interpretation rather than strict logical entailment, making it unsuitable for formal reasoning without modification (Figure~\ref{fig:dataset}).

\paragraph{Label Redefinition.}
We adopt a formal notion of entailment aligned with solver-based reasoning. A hypothesis $H$ is \textbf{entailed} by a premise $P$ if $P \wedge \neg H$ is unsatisfiable, and \textbf{contradicted} if $P \wedge H$ is unsatisfiable. All remaining cases are labeled \textbf{Neutral}.

Crucially, \textbf{Neutral} includes both (i) semantically irrelevant cases and (ii) cases where the premise is insufficient to support inference without additional assumptions.

\textbf{Example (Entailment) :} Under the premise \textit{``Nothing in this Agreement is to be construed as granting the Recipient any right whatsoever with respect to the Confidential Information,''} the hypothesis \textit{``Agreement shall not grant Receiving Party any right to Confidential Information''} is classified as 
\textbf{Entailment}, as $P \wedge \neg H$ is unsatisfiable.

\textbf{Example (Contradiction) :} Under the premise \textit{``Upon request, or if either party elects not to pursue any further business undertaking with the other, Recipient shall promptly return all tangible information, including any and all copies or partial copies thereof and thereupon confirm destruction of all information held electronically.''} the hypothesis \textit{``Receiving Party may retain some Confidential Information even after the return or destruction of Confidential Information.''} is classified as \textbf{Contradiction}, as $P \wedge H$ 
is unsatisfiable.

\textbf{Example (insufficient grounding):} Under the premise  
\textit{``The Recipient shall use the Confidential Information solely for the purpose for which it was disclosed,''}  
the hypothesis  
\textit{``Receiving Party shall not use any Confidential Information for any purpose other than the purposes stated in the Agreement''}  
is \textbf{Neutral}, as entailment requires the additional assumption that the disclosure purpose is defined in the Agreement.

\textbf{Example (irrelevance):} A hypothesis about retention of confidential information is \textbf{Neutral} under a premise governing third-party disclosure, as the premise is semantically unrelated to the hypothesis.
\paragraph{Data Annotation.}
\begin{table}[h!]
\centering
\renewcommand{\arraystretch}{1.3}
\begin{tabular}{lc}
\toprule
\textbf{Label Transition} & \textbf{Count} \\
\midrule
Entailment $\rightarrow$ Neutral     & 71 \\
Contradiction $\rightarrow$ Neutral  & 18 \\
Neutral $\rightarrow$ Entailment     & 14 \\
Neutral $\rightarrow$ Contradiction  &  4 \\
Entailment $\rightarrow$ Contradiction & 1 \\
Contradiction $\rightarrow$ Entailment  & 1 \\
\midrule
\textbf{Total Entailment}   & 153 \\
\textbf{Total Contradiction} & 52 \\
\textbf{Total Neutral}      & 295 \\
\bottomrule
\end{tabular}
\caption{Label transition statistics resulting from re-annotation under formal entailment definitions. The predominant shift from \textit{Entailment} and \textit{Contradiction} to \textit{Neutral} reflects the stricter logical grounding required for solver-based classification.}
\label{tab:label-transitions}
\end{table}

Using these definitions, we manually re-annotated 400 examples. We observed a substantial shift from \textit{Entailment} and \textit{Contradiction} to \textit{Neutral}, indicating that many originally labeled instances are not logically entailed under strict semantics. This shift reflects the prevalence of missing assumptions in real-world legal text.

Label transition statistics are reported in Table~\ref{tab:label-transitions}. The predominance of Entailment $\rightarrow$ Neutral transitions highlights a systematic gap between pragmatic interpretation and formal reasoning.

\paragraph{Dataset Augmentation.}
To further analyze this gap, we constructed minimal pairs for a subset of reclassified examples. For each instance labeled \textit{Neutral} due to missing information, we created a minimally modified hypothesis that becomes entailed or contradicted by explicitly supplying the required assumption.

\textbf{Example:} Under the premise  
\textit{``the undertakings apply to all information disclosed, regardless of the way or form in which it is disclosed or recorded,''}  
the hypothesis  
\textit{``Confidential Information may include verbally conveyed information''}  
is labeled \textbf{Neutral}, as equating verbal conveyance with \textit{form in which it is disclosed or recorded} requires an ungrounded assumption. The augmented hypothesis  
\textit{``Confidential Information may include recorded information''}  
is \textbf{Entailment}, as it is directly supported by the premise.

This construction isolates the role of implicit assumptions and provides controlled contrasts between semantically similar hypotheses that differ only in logical grounding, increasing the dataset size to 610 examples.

\paragraph{Dataset Label Annotation}
Table~\ref{tab:iaa-summary} reports inter-annotator agreement. While agreement is substantial (81.0\%, $\kappa = 0.627$), disagreements are concentrated between \textit{Entailment} and \textit{Neutral}. This indicates that the primary challenge lies not in logical inconsistency but in determining whether a hypothesis is sufficiently supported by the premise. Even among annotators with formal reasoning training, long or structurally complex sentences make label assignment difficult, particularly when it is unclear whether additional assumptions are required, reflecting the limits of deterministic logical interpretation under complex sentence structures. In cases of disagreement, a logician adjudicates the final label.
\begin{table}[ht]
\centering

\begin{tabular}{lc}
\toprule
\textbf{Metric} & \textbf{Value} \\
\midrule
Annotators & Annotator-1 \& Annotator-2 \\
Cohen's $\kappa$ & 0.627 \\
Interpretation (Landis \& Koch) & Substantial \\
Percent agreement & 81.0\%\\
Percent Disagreements & 19.0\% \\
\bottomrule
\end{tabular}
\caption{Inter-annotator agreement summary.}
\label{tab:iaa-summary}
\end{table}

\subsection{Neuro-symbolic Pipeline}
\begin{figure}[t]
\vskip 0.05in
\begin{center}
     \includegraphics[width=1\linewidth]{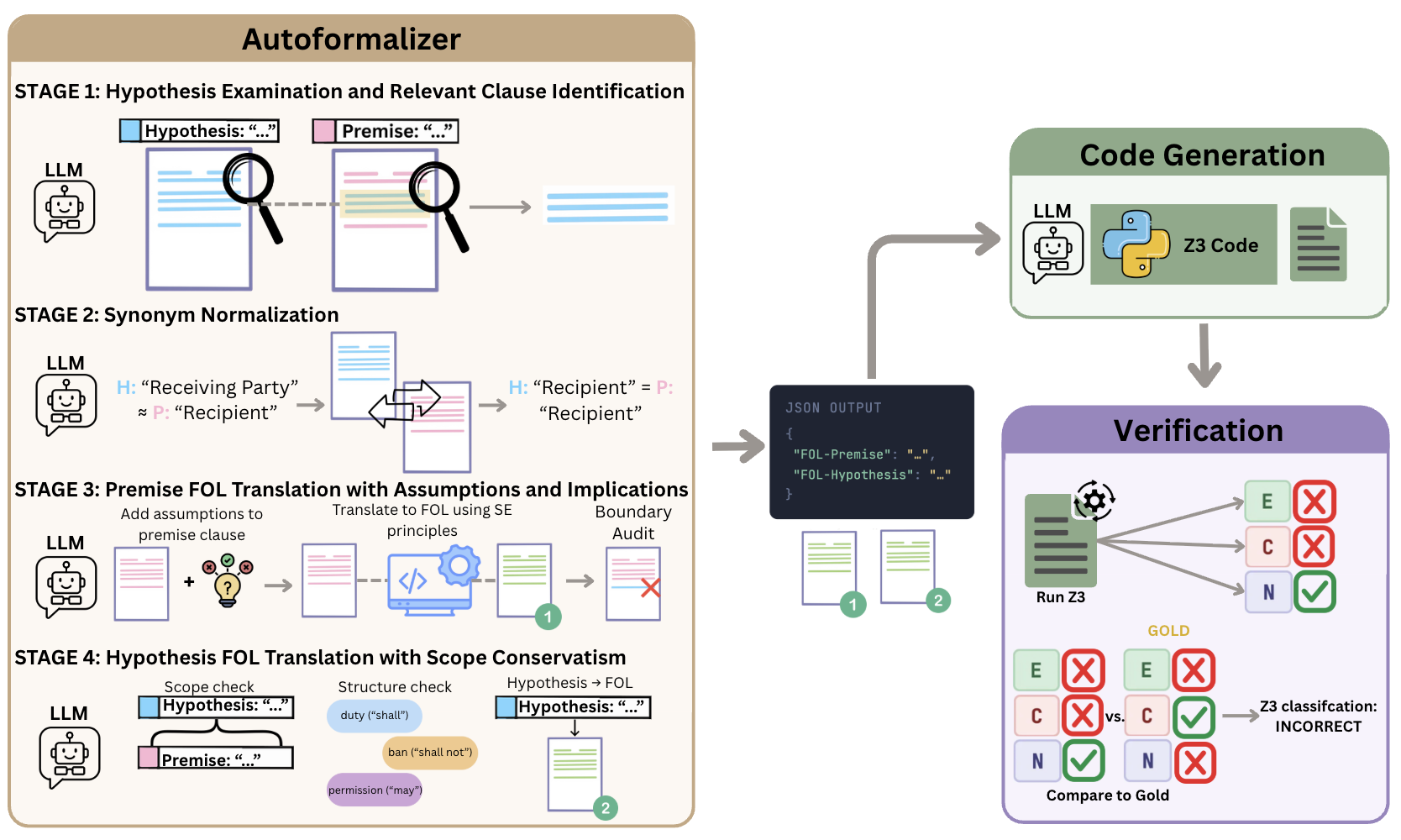}
     \caption{Neuro-symbolic pipeline for legal entailment. An LLM autoformalizes natural language into First-Order Logic (FOL), which is evaluated by the Z3 SMT solver for logically grounded classification. In parallel, the LLM performs native SMT reasoning over the same formal inputs. This separation enables analysis of errors from missing assumptions in formalization versus misinterpretation of logical representations.}
     \label{fig:pipeline}
\end{center}
\vskip -0.2in
\end{figure}

Our pipeline consists of two components: an \textbf{Autoformalizer}, which translates natural language into First-Order Logic (FOL), and a \textbf{Solver}, which performs entailment classification under formal semantics (Figure~\ref{fig:pipeline}). This design separates language understanding from logical reasoning, allowing us to isolate where failures occur.

\subsubsection{Autoformalizer}

The Autoformalizer uses a structured prompting workflow to generate FOL representations from legal text. The key stages are as follows:

\paragraph{(1) Relevant Clause Identification.}
Rather than translating the entire premise, the LLM first identifies clauses relevant to the hypothesis, following legal reasoning practices such as IRAC \citep{kang2023can}. This focuses formalization on semantically material content.

\paragraph{(2) Predicate Normalization.}
To ensure consistency in symbolic reasoning, the LLM aligns semantically equivalent expressions between premise and hypothesis, reducing errors caused by mismatched predicate representations.

\paragraph{(3) Formal Translation.}
The LLM translates both the premise and the hypothesis into FOL. Importantly, it is also prompted to encode assumptions in legal contract context as explicit logical statements. This step is critical, as formal reasoning requires all relevant assumptions to be explicitly represented.

\paragraph{(4) Global Constraints.}
We impose additional constraints to improve robustness, including prohibiting degenerate solutions (e.g., non-existence assumptions) and encouraging interpretable predicate construction.

\subsubsection{Solver}

Given the FOL representations, we use the Z3 SMT solver \citep{de2008proofs} to perform entailment classification based on satisfiability checks. This ensures that all predictions are logically valid with respect to the provided formalization.

To evaluate how well LLMs align with formal reasoning, we additionally prompt the same model to perform \textit{LLM-based formal reasoning and classification} directly over the SMT representations. Comparing these classification with solver outputs allows us to distinguish between errors arising from formalization versus those due to reasoning over structured representations.

\section{Results}
\subsection{LLM Experiment Results}
Table~\ref{tab:model_comparison} compares performance across methods. Introducing structure improves performance, with \textit{LLM-based Formal Reasoning} achieving the highest accuracy for several models. In contrast, solver-based classification is the most conservative, reflecting strict adherence to formal semantics.
Table~\ref{tab:error_analysis} reveals distinct error patterns:

\paragraph{LLM Over-Entailment.}
Pure LLM methods (\textit{Simple}, \textit{Structured}) frequently misclassify \textit{Neutral} cases as \textit{Entailment} (high N$\rightarrow$E rates), indicating reliance on implicit assumptions.

\paragraph{Formal Reasoning as Intermediate.}
LLM reasoning over FOL reduces some over-prediction while maintaining strong accuracy, but still diverges from solver outputs, suggesting incomplete adherence to formal semantics.

\paragraph{Solver Conservativeness.}
Solver-based classification yields notably more conservative results than LLM-based classification, with a substantial proportion of instances labeled Neutral due to insufficient information. Detailed results and analysis are provided in the Appendix Table \ref{tab:smt-performance}.

\paragraph{Error Asymmetry.}
Across models, confusion between \textit{Neutral} and \textit{Entailment} dominates, while \textit{Entailment}–\textit{Contradiction} errors are rare, indicating that the primary difficulty lies in insufficient information rather than logical inconsistency.

\begin{table}[h!]
\centering
\begin{tabular}{l c c c c }
\hline
\textbf{Model} & \textbf{Simple} & \textbf{Structured} & \textbf{LLM-FR} & \textbf{Solver-FR} \\
\hline
Claude   & 63.1\% & 53.2\% & 83.0\% & 74.5\% \\ 
Deepseek & 68.4\% & 62.2\% & 64.2\% & 45.3\% \\
GPT      & 66.0\% & 58.8\% & 65.3\% & 60.3\%  \\
Qwen      & 65.1\% & 61.5\% & 69.1\% & 48.3\% \\
Llama    & 55.9\% & 40.0\% & 58.6\% & 42.1\% \\

\hline
\end{tabular}
\caption{Performance Comparison across Classification Approaches. All values report accuracy (\%). \textit{Simple}: pure LLM
classification using the annotation guideline as the prompt. \textit{Structured}: pure LLM classification with the same prompt as LLM-Formal Reasoning, but direction classification as outputs.
\textit{LLM-FR}: LLM-based formal reasoning classification where the premise and hypothesis are autoformalized into First-Order Logic, and the LLM is prompted to produce a classification by running the solver. \textit{Solver-FR}: Z3 solver-based formal reasoning classification where the autoformalized premise and hypothesis are run through the Z3 solver for the classification results. \textit{Baseline}: We define baseline performance as the proportion of the majority class, equivalent to the accuracy of a majority-class classifier, which is 48\%}
\label{tab:model_comparison}
\end{table}

\begin{table}[h!]
\centering
\begin{adjustbox}{max width=\textwidth}
\begin{tabular}{l l c c c c c c}
\hline
\textbf{Model} & \textbf{Method} & \textbf{N$\to$E (\%)} & \textbf{E$\to$N (\%)} & \textbf{N$\to$C (\%)} & \textbf{C$\to$N (\%)} & \textbf{C$\to$E } & \textbf{E$\to$C (\%)} \\
\hline
\multirow{4}{*}{Claude}
 & Simple          & 20.3\%  & 2.5\% & 10.5\%  & 0.5\% & 0.3\%  & 2.5\%  \\
 & Structured      & 28.9\%   & 0.4\%   & 15.0\%   & 0   & 0.5\%   & 1.8\%   \\
 & LLM-Formal & 8.9\%  & 6.5\%   & 2.0\%   & 1.3\%   & 0.7\%   & 0.4\%   \\
 & Z3-Formal & 3.9\%  & 5.4\%   & 0.7\%   & 1.8\%   & 0.2\%   & 0\%   \\
\hline
\multirow{4}{*}{Deepseek}
 & Simple          & 12.1\%  & 9.8\%  & 4.9\%  & 4.3\%  & 0.18\% & 0.18\%   \\
 & Structured      & 23.7\%   & 2.4\%   & 8.5\%   & 2.0\%  & 0.5\% & 0.4\%  \\
 & LLM-Formal & 2.9\% & 26.2\%   & 2.2\%   & 14.1\%   & 0   & 0   \\
 & Z3-Formal & 1.1\%  & 21.5\%   & 0.8\%   & 7.4\%   & 0.3\%   & 0.2\%   \\
\hline
\multirow{4}{*}{ChatGPT}
 & Simple          & 17.4\% & 6.5\% & 7.8\%  & 1.1\%  & 0.4\% & 0.7\%   \\
 & Structured      & 20.4\%   & 5.2\%   & 10.3\%   & 0   & 0.5\%   & 1.8\%   \\
 & LLM-Formal & 7.4\% & 17.9\%  & 0.4\%  & 6.9\%  & 0.5\% & 0  \\
 & Z3-Formal & 8.2\%  & 18.1\%   & 0\%   & 7.1\%   & 0.2\%   & 0.5\%   \\
\hline
\multirow{4}{*}{Qwen}
 & Simple          & 25.9\% & 1.3\% & 4.2\% & 2.9\% & 0.5\% & 0  \\
 & Structured      & 26.2\% & 1.1\% & 8.7\% & 1.6\% & 0.5\% & 0.2\% \\
 & LLM-Formal & 15.4\%  & 9.6\%   & 2.4\%   & 4.7\%   & 0.4\%   & 0.5\%   \\
 & Z3-Formal & 12.6\%  & 14.9\%   & 2.3\%   & 5.1\%   & 0.3\%   & 0.8\%   \\
\hline
\multirow{4}{*}{Llama}
 & Simple          & 30.4\%   & 4.2\%   & 4.9\%   & 4.0\%   & 0.4\%   & 0.2\%    \\
 & Structured      & 42.9\%   & 2.4\%   & 12.1\%   & 1.6\%   & 0.2\%   & 0.7\%   \\
 & LLM-Formal & 18.8\%  & 6.7\%   & 3.4\%   & 2.5\%   & 1.4\%   & 2.2\%   \\
 & Z3-Formal & 11.3\%  & 27.3\%   & 3.9\%   & 5.4\%   & 2.3\%   & 0.8\%   \\
\hline
\end{tabular}
\end{adjustbox}
\caption{Classification Error Analysis across Models and Methods. Each cell reports the percentage of samples misclassified in the given direction, where N = Neutral, E = Entailment, C = Contradiction, and X$\to$Y denotes samples with true label X
predicted as Y.}
\label{tab:error_analysis}
\end{table}

\section{Discussion}

\subsection{Ambiguity in Permissible Assumptions}
Our results consistently point to a central challenge: the difficulty is that it is fundamentally unclear which assumptions are permissible in legal reasoning, and LLMs frequently introduce ungrounded ones.

This ambiguity is evident in both human annotation and model behavior. Inter-annotator disagreements are dominated by \textit{Entailment}–\textit{Neutral} confusion, indicating that even humans struggle to determine whether a hypothesis is sufficiently supported by the premise, when the sentence gets lengthy and complex. Similarly, across all models and methods, the dominant error mode is \textit{Neutral} $\rightarrow$ \textit{Entailment}, suggesting that LLMs systematically introduce implicit assumptions to bridge missing information. In contrast, solver-based formal reasoning exhibits the opposite tendency, frequently classifying cases as \textit{Neutral} due to missing explicit constraints. This divergence highlights a fundamental tension: LLMs resolve ambiguity through assumption injection, while formal methods expose it through conservative reasoning.

\subsection{Accuracy–Faithfulness Trade-off}

We observe a consistent trade-off between accuracy and logical faithfulness. LLM-based methods, particularly those reasoning over formal representations, achieve higher accuracy by leveraging implicit or unverified assumptions. However, these assumptions are not grounded in the premise.

By contrast, solver-based reasoning enforces strict logical validity and therefore produces more conservative outputs. In many cases, hypotheses labeled as \textit{Entailment} by LLMs are classified as \textit{Neutral} by the solver because the required bridging assumptions are absent from the formal representation.

\subsection{Limits of Autoformalization}

Our experiments show that autoformalization is the primary bottleneck in neuro-symbolic systems. Even with structured prompting and explicit instructions to surface assumptions, LLM-generated formalizations remain incomplete or incorrect.

A representative example is shown below. For the premise:

\textit{``If requested, either Party shall be bound to return any and all materials to the Requesting Party within \_ days.''}

and the hypothesis:

\textit{``Receiving Party may retain some Confidential Information even after the return or destruction of Confidential Information.''}

The autoformalized FOL introduces multiple auxiliary axioms, including permissions, return obligations, and survival clauses, which are not grounded in the premise. 

This example highlights a key limitation: autoformalization can introduce ungrounded assumptions, blurring the boundary between valid inference and hallucination. More broadly, the challenge lies not only in translating language into formal representations, but in determining which implicit assumptions should be made explicit. Even with explicit prompting, LLMs fail to consistently recover the minimal assumptions required for faithful reasoning.

\subsection{Error Patterns}
These results reveal that errors in our pipeline are not random, but fall into a small number of recurring and interpretable failure modes.

\paragraph{Translation Error Analysis.}
We observe hallucinated axioms, such as survival obligations or harm assumptions, are introduced without grounding in the source text. The prevalence of such translation errors underscores that the autoformalization process can actively introduce incorrect assumptions.

\paragraph{Scope Laundering Failure.}
We observe a distinct and recurring failure mode in which LLMs report solver-consistent outputs without executing the underlying code. In these cases, the model produces classifications such as \textit{Entailment} or \textit{Contradiction}, while actual solver execution yields \textit{Neutral}. This indicates that the model is reasoning informally and presenting its output as if derived from symbolic execution. We refer to this as \textbf{scope laundering}, where the hypothesis appears logically grounded when it is not, and identify it as one of the more prevalent failure modes in our analysis. All models exhibit the patterns of scope laundering, varying from as low as 15.3\% in GPT and as high as 52.5\% in Qwen (28.6\% if invalid inputs are excluded). These astonishing results raise serious questions about the faithfulness of in "LLM-based formal reasoning". Even though their performances may look better, it may come at the cost of faithfulness. 

\paragraph{Implicit Constraint Blindness.}
Across models, we observe a systematic pattern whereby the SMT solver produces correct non-neutral classifications while LLM-based reasoning fails. Analysis reveals that LLMs consistently struggle to capture implicit logical constraints, such as universals encoded 
within existential structures. We term this failure mode \textbf{implicit constraint blindness}, where the model overlooks constraints necessary for correct inference despite their presence in the formal representation. We also observed this error pattern across models, varying from as low as 0.7\% in GPT and as high as 4.4\% in Claude. 

\paragraph{Program Synthesis Failures.}
We observe substantial variation in the quality of LLM-generated Z3 Python code, indicating that autoformalization also involves a non-trivial program synthesis challenge. Claude performs best, with errors in roughly 25.5\% of cases, mainly sort mismatches and occasional syntax issues. Llama performs poorly, often mixing SMT-LIB with Python or failing to produce executable code, with errors including misuse of strings as sorts, malformed quantifiers, and type inconsistencies. DeepSeek, GPT, and Qwen show moderate performance but still require frequent correction, with common issues such as sort mismatches, variable scoping errors, quantifier misuse, and type inconsistencies. Detailed results and analysis are provided in the Appendix table \ref{tab:smt-performance}. 

\subsection{Limitations}
Our study has several limitations. The dataset is relatively small and drawn from a single source, with an imbalanced label distribution. Annotation remains subjective in edge cases, particularly when distinguishing between "entailment" and "neutral". Additionally, some autoformalization outputs remained invalid even after our three-iteration pipeline and required manual correction due to execution errors.

Our findings suggest that the core challenge lies not in the choice of formalism but in constructing representations that capture all relevant assumptions. Even under a strictly logical framework, recovering the implicit assumptions necessary to generate executable formal representations remains non-trivial. LLMs in particular struggle to reliably recover the implicit knowledge required for correct reasoning, suggesting that robust assumption and ambiguity handling is a prerequisite for more expressive logical systems.

Despite these limitations, our goal is not to optimize benchmark performance but to examine how different reasoning modes diverge under strict logical semantics on long, complex sentences such as those found in legal text. Expanding this analysis to larger and more diverse datasets remains an important direction for future work.

\subsection{Future Work}

\paragraph{Improving Autoformalization.}
Since autoformalization is the primary bottleneck, improving representation quality is critical. Synonym normalization could be strengthened using embeddings or legal ontologies to canonicalize entities. Pipeline decomposition into specialized agents (e.g., clause selection, predicate design, translation, verification) may improve robustness. Extending to document-level reasoning would enable cross-clause dependencies, while richer formalisms such as deontic logic could better capture normative distinctions currently collapsed in FOL.

\paragraph{Assumptions and Ambiguity in Legal Reasoning.}
Legal reasoning depends on background assumptions and contextual interpretation \citep{sartor2005legal, rotolo2015deontic}. For example, contractual obligations are typically understood as excluding illegal conduct, even when unstated, but must be explicitly encoded in formal systems. This mismatch reflects broader concerns in computational law about the loss of interpretive flexibility when legal text is reduced to formal representations \citep{hildebrandt2018law, diver2021interpreting}. A promising direction is to integrate structured legal knowledge (e.g., ontologies, rule-based systems) with LLMs that propose context-dependent assumptions during formalization \citep{francesconi2023patterns}. We propose a complementary and more targeted approach: surfacing Minimal Correction Subsets (MCS) via SMT solvers \citep{singh2026verge} and presenting them to legal practitioners as structured entry points for resolving ambiguity. Rather than asking lawyers to specify all relevant background knowledge upfront, this approach begins from the conservatively verified facts established by the solver and identifies the minimal set of axioms whose acceptance would shift the classification from Neutral to Entailment or Contradiction. This reframes legal ambiguity not as a failure of the system but as an inherent property of legal text that AI alone cannot resolve, and positions the lawyer as the appropriate decision-maker for the specific, well-scoped interpretive questions the system cannot answer. The result is a lawyer-centered reasoning tool that constrains human review to precisely the assumptions that matter, rather than requiring exhaustive document-level verification.

\paragraph{Learning for Faithful Formalization.}
While recent work has explored training LLMs for improved logical reasoning and formal consistency, the models evaluated in this work are general-purpose systems not specifically optimized for legal autoformalization. Our results suggest that this setting introduces additional challenges beyond standard formal reasoning, including domain-specific language, assumptions, and strict requirements for logical grounding. A promising direction is to incorporate solver-based feedback, such as satisfiability signals, into training objectives tailored to legal reasoning tasks. Our error analysis highlights issues such as predicate inconsistency, hallucinated axioms, and missing constraints. This provides a foundation for constructing targeted training curricula and evaluation benchmarks specific to legal autoformalization.

\paragraph{Beyond Legal Domain.}
This framework extends to domains such as financial regulation and cybersecurity compliance \citep{tumkur2025neuro, hsia2025neuro, zhang2025neuro}, where reasoning similarly depends on making implicit assumptions explicit for reliable formal verification.

\section{Conclusion}
This work presents a systematic study of LLM-based and neuro-symbolic reasoning on real-world legal text, revealing a fundamental gap between benchmark accuracy and logical faithfulness. Our re-annotation of ContractNLI under strict formal semantics demonstrates that this gap originates in the data itself: a substantial proportion of legally sound inferences depend on implicit assumptions absent from the contract text, creating a systematic divergence between pragmatic legal interpretation and formal entailment that neither humans nor models consistently resolve. While formal structure improves benchmark performance, it does not imply faithful reasoning. Scope laundering across board rreveals that LLMs frequently produce classifications that appear formally grounded without executing the underlying symbolic reasoning. This fundamentally undermines the use of LLM-based formal reasoning as a faithful proxy for solver-based verification. Implicit constraint blindness and program synthesis failures further show that even correct formal representations are not reliably applied or generated by current models. The absence of a clear boundary between valid inference and unjustified assumption is an ambiguity that affects human annotation and model behavior alike. Even under a strictly logical annotation framework, the length and structural complexity of legal language make the deterministic classification task difficult for human annotators while LLMs struggle further, often failing to produce executable representations that are both correct and semantically complete. Progress in faithful legal AI will require not only better models, but methods that make this boundary explicit and actionable, surfacing the minimal assumptions underlying each inference for targeted human review rather than requiring exhaustive verification.

\newpage
\section*{Ethics Statement}
The application of Large Language Models (LLMs) to legal reasoning introduces significant ethical considerations regarding professional responsibility, accuracy, and systemic bias. In this work, we propose a neuro-symbolic framework designed to mitigate the risks of LLM hallucination by introducing a deterministic symbolic reasoning layer. However, the following ethical dimensions must be considered:

\paragraph{Human-in-the-Loop Necessity: } While our pipeline improves the verifiability of legal entailment, it is not intended to replace human legal counsel. The outputs of the SMT solver and the LLM-generated formalizations should be treated as decision-support tools, not as binding legal advice or autonomous compliance certifications.

\paragraph{Risks of Automation Bias: } There is a risk that users may over-rely on the "formal" nature of symbolic outputs, assuming they are infallible. We emphasize that the fidelity of the reasoning is strictly dependent on the quality of the autoformalization stage, which remains susceptible to translation errors.

\paragraph{Dataset and Annotation Bias: } Our findings are grounded in a re-annotated subset of ContractNLI, which reflects the specific interpretations and educational backgrounds of our annotators. These interpretations may not align with all legal jurisdictions or the specific intent of different contracting parties.

\paragraph{Interpretability and Accountability: } By using a neuro-symbolic architecture, we aim to provide an auditable "chain of logic" that allows human reviewers to inspect why a specific classification was reached. This transparency is a deliberate choice to support accountability in high-stakes legal deployments.

\paragraph{Dual-Use and Accessibility:} While this technology can assist in contract due diligence, it could also be used to automate the identification of legal loopholes for exploitative purposes. We advocate for the responsible deployment of these tools within established legal and regulatory frameworks.

\newpage
\bibliography{colm2026_conference}
\bibliographystyle{colm2026_conference}
\newpage
\appendix
\section{Appendix}
\subsection{SMT Solver Experiment Results}
Due to the high error rate in the generated Python code, we created a code-fixing pipeline and fed the erroneous code, along with the error message, back to the respective model and asked it to fix the code. We did this for three iterations, and if, after three iterations, the code was still not fixed, we marked it as "Invalid". If the code was fixed, we executed it to obtain classification results.
\textbf{Invalid} captures responses that either returned no code, produced unparseable output, or generated code that failed the premise sanity check by yielding a contradictory result.
\textbf{Error Rate} measures the overall percentage of incorrect classifications per model. We report the classification errors and invalid code percentage in table \ref{tab:smt-performance}
These results show that reliable formalization requires not only correct semantics but also executable representations, which current LLMs struggle to produce. However, more exposure to the specific formal language could help decrease the error rates significantly. 

\begin{table}[ht]
\centering
\resizebox{0.4\textwidth}{!}{%
\begin{tabular}{lcc}
\toprule
\textbf{Model} & \textbf{Error Rate} & \textbf{Invalid (\%)} \\
\midrule
Claude   & 25.5\% & 0 \\
DeepSeek & 54.7\% & 23.6\% \\
GPT      & 39.7\% & 6.1\% \\
Qwen     & 51.7\% & 15.6\% \\
Llama    & 63.2\% & 28.7\% \\
\bottomrule
\end{tabular}%
}
\caption{SMT solver solving performance across models.}
\label{tab:smt-performance}
\end{table}
Overall our findings showed a big gap among different models' capability of fixing their own code. However, we do not claim that this applies to all different types of code, as our situation is limited to Z3 python code that follows a specific format. We attribute the perfect "invalid" percentage from Claude to a much more significant exposure to Z3 python code compared to other models. 
\subsection{Annotation Guideline \& Simple Prompt for Pure LLM Classification}
Annotation Guidelines for Contractual Entailment
1. Annotation Objective
Annotators must determine whether a hypothesis is logically supported by the premise.
Each example should be labeled as one of:
Entailment, Contradiction, or Neutral.

The goal is to evaluate strict textual entailment, not general plausibility or likely interpretation.
Annotators should rely only on the premise and should not assume additional legal knowledge or common business practices.

2. Label Definitions
Entailment
A hypothesis is Entailed if the premise explicitly guarantees that the hypothesis must be true.
This means:
1. The hypothesis follows directly from the premise
2. No assumptions beyond the premise are required.
3. The hypothesis would always hold whenever the premise is valid.

Key rule
If the premise is true, the hypothesis must also be true.
Examples
Premise : “The recipient shall not disclose confidential information to any third party.”
Hypothesis : “The recipient is prohibited from disclosing confidential information to third parties.”
Label : Entailment

Premise : “The agreement may be terminated by either party with 30 days written notice.”
Hypothesis : “Either party can terminate the agreement by providing 30 days written notice.”
Label : Entailment

Premise : “The service provider will maintain the system uptime at 99.9
Hypothesis : “The provider is obligated to maintain system uptime at 99.9
Label : Entailment

Entailment checklist
Annotators should confirm:
1. The hypothesis is directly supported by the premise
2. The hypothesis does not extend or generalize beyond reasonable implications from the premise

3. Contradiction
A hypothesis is Contradicted if the premise explicitly states the opposite or makes the hypothesis impossible.
This means:
1. The hypothesis conflicts with the premise.
2. If the premise is true, the hypothesis cannot be true.

Key rule
The premise logically rules out the hypothesis.
Examples
Premise : “The recipient shall not disclose confidential information to third parties.”
Hypothesis : “The recipient may disclose confidential information to third parties.”
Label : Contradiction

Premise : “The agreement may be terminated only by the service provider.”
Hypothesis : “Either party may terminate the agreement.”
Label : Contradiction

Premise : “The company must provide written notice before terminating the agreement.”
Hypothesis : “The company may terminate the agreement without written notice.”
Label : Contradiction

Contradiction checklist
Annotators should confirm:
1. The hypothesis conflicts with explicit premise language
2. Both the premise and the hypothesis cannot be true simultaneously

4. Neutral
A hypothesis is Neutral if the contract text does not provide enough information to determine whether the hypothesis is true or false.
This occurs when:
1. The premise does not mention the topic.
2. The hypothesis is broader than the premise clause.
3. The hypothesis requires external knowledge or is beyond reasonable assumptions.

Key rule
The hypothesis may or may not be true, based on the text.

Examples
Premise : “The recipient shall not disclose confidential information to third parties.”
Hypothesis : “The recipient may store confidential information securely.”
Label : Neutral

Premise : “The agreement may be terminated with 30 days written notice.”
Hypothesis : “The agreement may be terminated by the company.”
Label : Neutral
Reason: The clause does not specify who may terminate.

Premise : “The service provider will maintain uptime of 99.9\%.”
Hypothesis : “The service provider guarantees uninterrupted service.”
Label : Neutral
Reason: 99.9\% uptime does not imply uninterrupted service.

Neutral checklist
Annotators should check whether the hypothesis:
1. introduces new entities or conditions
2. generalizes beyond the clause
3. requires world knowledge
4. is not addressed by the contract

5. Common Annotation Pitfalls
Annotators should avoid the following mistakes.
1. Inferring typical business practices
Premise : “The provider may terminate the agreement.”
Hypothesis : “The provider must provide advance notice.”
Label : Neutral
Reasoning : Even if advance notice is typical, the premise does not state it.

2. Over-generalizing from a specific clause
Premise : “The company may disclose confidential information to affiliates.”
Hypothesis : “The company may disclose confidential information.”
Label : Neutral
Reasoning : The hypothesis removes the affiliate restriction.

3. Treating permissions as obligations
Premise : “The company may disclose confidential information.”
Hypothesis : “The company must disclose confidential information.”
Label : Neutral
Reasoning : Permission does not imply obligation.
The hypothesis is not a CONTRADICTION because it is logically consistent with the premises — it does not conflict with them, it merely goes further than they establish (think of it as the hypothesis has a bigger scope than the premise, but in the same direction).

6. Modal Language Interpretation
Contracts often contain modal verbs. Annotators should interpret them carefully.
Modal phrase -> Interpretation
shall / must -> obligation
may -> permission
shall not / may not -> prohibition
only if -> conditional constraint

Examples
Premise : “The company may disclose information to affiliates.”
Hypothesis : “The company must disclose information to affiliates.”
Label : Neutral

Premise : “The company must disclose information to affiliates.”
Hypothesis : “The company may disclose information to affiliates.”
Label : Entailment

7. Decision Process
Annotators should follow this decision order:
Does the premise explicitly imply the hypothesis? $\rightarrow $ Entailment

Does the premise explicitly rule out the hypothesis? $\rightarrow $ Contradiction

Otherwise $\rightarrow $ Neutral

8. Annotation Confidence
Annotators should flag examples (classify as “neutral” and highlight the row in red, add your reasoning in col D) that are:

1. ambiguous

2. poorly worded

3. dependent on missing definitions

These examples can be reviewed separately.

\end{document}